\documentclass[11pt, a4paper]{custom_class}

\usepackage{custom_style}
\usepackage[utf8]{inputenc} 
\usepackage[T1]{fontenc}    
\usepackage{hyperref}       
\usepackage{url}            
\usepackage{booktabs}       
\usepackage{amsfonts}       
\usepackage{nicefrac}       
\usepackage{microtype}      
\usepackage{cleveref}       
\usepackage{lipsum}         
\usepackage{graphicx}
\usepackage{natbib}
\usepackage{doi}
\usepackage{natbib}
\title{Anthropocentric bias\\in language model evaluation}

\author{
Raphaël Millière \\
University of Oxford \\
\texttt{raphael.milliere@philosophy.ox.ac.uk} \\
\And
Charles Rathkopf \\
Forschungszentrum Jülich \\
\texttt{c.rathkopf@fz-juelich.de} \\
}

\begin{document}
\maketitle

\begin{center}
  \textcolor{red}{Published in \textit{Computational Linguistics}}\\[0.5em]
  \href{https://doi.org/10.1162/COLI.a.582}{\texttt{https://doi.org/10.1162/COLI.a.582}}
\end{center}

\begin{abstract}
Evaluating the cognitive capacities of large language models (LLMs) requires overcoming not only anthropomorphic but also anthropocentric biases. This article identifies two types of anthropocentric bias that have been neglected: overlooking how auxiliary factors can impede LLM performance despite competence (\textit{auxiliary oversight}), and dismissing LLM mechanistic strategies that differ from those of humans as not genuinely competent (\textit{mechanistic chauvinism}). Mitigating these biases requires an empirical, iterative approach to mapping cognitive tasks to LLM-specific capacities and mechanisms, achieved by supplementing behavioral experiments with mechanistic studies.
\end{abstract}

\section{Introduction}
\label{submission}

What cognitive competencies do large language models (LLMs) have, if any? That is a question of immense theoretical and practical importance. It is also formidable, as we lack an appropriate methodological framework to answer it. The rigorous methodology of experimental psychology (e.g., controlled experiments, counterbalanced stimuli) provides an excellent foundation for investigating the cognitive capacities of LLMs. However, we cannot transfer its paradigms whole cloth, since inferences from these paradigms in human studies rest on background assumptions about cognitive architecture that may not hold for LLMs.\footnote{See \citet{ivanovaHowEvaluateCognitive2025} for a related discussion.} As a result, naively applying these methods to LLMs risks succumbing to both anthropomorphic and anthropocentric biases. Anthropomorphic bias involves attributing human qualities to LLMs without justification – being too eager to recognize the capacities of an LLM as instances of our own. Anthropocentric bias, conversely, entails evaluating LLMs by human standards without justification, and refusing to acknowledge genuine cognitive competence that differs in substantive ways from our own. We aim to clarify this challenge, offering a path forward for resolving disputes and advancing our understanding of LLMs' capacities.\footnote{We do not aim to discuss whether LLMs \textit{should} be designed to have human-like capacities, but how to fairly assess and compare their capacities to human cognition.} 

\section {The performance/competence distinction}

The distinction between competence and performance \citep{chomskyAspectsTheorySyntax1965} plays a crucial role in cognitive science. Competence if often defined as the system's internal knowledge underlying a particular capacity, and performance as the observable behavior of a system exercising it \citep{firestonePerformanceVsCompetence2020}. For a competent system, then, performance is the external manifestation of competence. This formulation allows for cases where performance and competence diverge. For instance, a student cheating on a test demonstrates performance success without competence, while a knowledgeable student failing due to anxiety shows performance failure despite competence. Such examples might invite criticism that applying this distinction to LLMs is inherently anthropomorphic, by relying on a concept of knowledge proper to humans. We can avoid this concern with an alternative formulation. In an experimental context, \textit{performance} refers to how closely a system's behavior aligns with some normative standard of success on a task, while \textit{competence} refers to system's computational capacity to meet that normative standard under fair testing conditions.\footnote{This is meant to exclude ideal conditions where a system simply gets ``lucky''---for example, where the correct answers in a test happen to be high-frequency tokens. Rather, fair testing conditions are those where cognitive mechanisms for the core competence can operate without impediment from auxiliary factors.} 

It is widely recognized that there is a double dissociation between performance and competence, and that this distinction gives us reason to be wary not only of naive inferences from good performance to competence, but also from bad performance to lack of competence. Although inferences about competence from performance are drawn in both directions in human experimental psychology, they are drawn in only one direction  when studying LLMs: from bad performance to lack of competence. Without adequate justification, this asymmetry is suggestive of anthropocentric bias. We suspect that this reflects the reasonable desire to temper the opposite bias – \textit{anthropomorphism}, the tendency to ascribe human-like capacities to LLMs without sufficient evidence.\footnote{For an example of the strong anti-anthropomorphism sentiment, which we suggest may motivate this counter-reaction, see \citet{benderNYTMagazineAI2022}.} However, fighting one bias by entrenching another won't foster impartiality  \citep{bucknerBlackBoxesUnflattering2021}. Instead, as the philosopher Elliot Sober once remarked in a discussion of anthropocentric reasoning in comparative psychology, ``the only prophylactic we need is empiricism'' \citep[][p. 97]{soberComparativePsychologyMeets2005}.

\section{Taxonomy of Anthropocentric Bias}

A pattern of reasoning is biased when it lacks adequate justification. Since justification can fail in different ways, each failure corresponds to a distinct variety of anthropocentric bias with its own methodological challenges. We outline a taxonomy of these biases and strategies to mitigate them.

\subsection{Auxiliary oversight}

The first kind of bias, which we call \textit{auxiliary oversight}, is the tendency to overlook the persistent possibility that an LLM's performance failures on a task designed to measure competence $C$ may be caused by \textit{auxiliary factors} rather the lack of competence $C$. This bias arises from a failure to account for the different cognitive profiles of LLMs and humans, which can lead to flawed experimental designs in two ways. First, by providing LLMs with less support than humans for the same task, creating an unfair comparison. Second, by designing tasks with components that are trivial for humans but non-trivial for LLMs due to architectural differences. In both cases, the assumption is flawed because it overlooks the possibility that auxiliary factors caused the performance failure.

In human psychology, auxiliary factors are often illustrated by cases such as the student described above: the student's nervousness negatively impacts her performance on the test, even though she has the relevant knowledge and would have otherwise done well. Applying this concept to LLMs might seem anthropomorphic, as it relies in the human context on the existence of a complex, semi-modular faculty psychology, where limitations in one faculty might bottleneck performance in another. Since LLMs presumably do not have psychological faculties in that sense, citing auxiliary factors as causes of LLM performance failure may seem like a nonstarter.

However, auxiliary factors can and do influence LLM performance. To understand their role, it is helpful to recall that every mechanistic explanation involves a degree of idealization: it picks out a subset of the many causal influences on behavior. In doing so, it implicitly assumes that the remaining factors will either remain stable or exert only minor influence. When that assumption fails, performance may degrade in ways that do not reflect a lack of underlying competence. Sometimes, failure results from a missing enabler -- a subsystem or context the subject relies on but does not receive. For example, in the mirror test for self-recognition, an animal may fail not because it lacks the capacity for self-representation, but because it lacks motivation to remove a visible mark. In other cases, failure stems from interference by competing mechanisms, as when a bilingual child struggles with a vocabulary test in one language due to lexical interference from the other. Finally, performance may be limited by how much processing the subject is permitted or able to perform -- as when a student asked to solve a math problem mentally fails, but succeeds with pen and paper. These scenarios violate the idealized assumptions built into our explanatory models. Auxiliary factors are not part of the competence being tested, yet their influence can mask or suppress it. Ignoring them risks conflating failure of performance with failure of cognition.

In LLMs, we can distinguish at least three kinds of auxiliary factors. The first and most familiar kind are auxiliary task demands. \citet{huAuxiliaryTaskDemands2024} provide a helpful illustration of such demands in evaluating whether language models are sensitive to syntactic features like subject-verb agreement. They compare two approaches: (1) prompting the model to make explicit grammaticality judgments, and (2) directly comparing the probabilities the model assigns to minimal pairs that vary the target feature. For the metalinguistic approach, they use prompts such as: ``Here are two English sentences: 1) Every child has studied. 2) Every child have studied. Which sentence is a better English sentence? Respond with either 1 or 2 as your answer.'' They then analyze both the output and the probabilities assigned to `1' and `2'. For the direct estimation approach, they prompt LMs with minimal pairs such as ``every child has studied'' and ``every child have studied'', and compare the log probabilities assigned to each string. A model is considered successful if it assigns a higher probability to the grammatical sentence. Across model sizes and datasets, Hu and Frank find that direct probability estimation yields results that differ from and are often better than the metalinguistic approach. They conclude that metalinguistic prompting introduces an auxiliary task demand - the ability to generate explicit grammaticality judgments - that is irrelevant to the underlying syntactic competence of interest. In contrast, direct probability estimation more validly measures the target capacity. We concur with their assessment. What makes it a genuine \textit{demand} is the fact it degrades performance. What makes it genuinely \textit{auxiliary} is the fact that metalinguistic judgment is conceptually independent of the psychological construct of interest, which is the capacity to track grammaticality.

Neglecting auxiliary task demands can lead inferences about competence astray. Such negligence is compounded in comparative studies with mismatched experimental conditions, resulting in divergent auxiliary task demands for LLMs and human subjects. This concern is highlighted by \citet{lampinenCanLanguageModels2023}'s case study comparing human and model performance on recursively nested grammatical structures, in response to prior work by \citet{lakretzCanTransformersProcess2022}. Lakretz et al. found that humans outperformed language models on challenging long-distance subject-verb agreement dependencies in embedded clauses. However, Lampinen notes that human subjects were given substantial instructions, training and feedback to orient them to the experimental task, while models were evaluated ``zero-shot'' without any task-specific context. The discrepancy in experimental conditions confounds the comparison: the additional context provided to humans but not models can be interpreted as imposing weaker auxiliary task demands on humans than models. To level the playing field, Lampinen tested LLMs on the same task by providing it with prompts containing a few examples, intended to match the orienting context given to human subjects. With this modest task-specific context, LLMs perform as well as or better than humans, even on challenging sentences with deeper nesting than those tested in humans. In a similar vein, \citet{huLanguageModelsAlign2024a} re-examined claims that LLMs fail at basic grammaticality judgments and found that apparent failures were an artifact of an experimental design that conflated linguistic competence with the auxiliary demand of metalinguistic reasoning. Once this demand was removed in favor of a more direct probabilistic evaluation, the models' performance was aligned with that of human subjects. These cautionary tales illustrate how mismatched experimental conditions across humans and models -- with respect to instructions, examples, motivation, and other factors -- can distort comparisons of their capacities. Meaningful comparative evaluation requires that humans and models are subject to similar auxiliary task demands, just as comparative psychology strives for ``species-fair comparisons'' across humans and animals.

Auxiliary task demands are challenging in LLM evaluation because tasks that are considered trivial for humans may not be trivial for an LLM. The classification of a task as trivial depends on whether the cause of performance failure is associated with the parts and operations that explain its success when things go well. If the cause of performance failure is related to \textit{other} parts of the system that do not contribute to its success in normal circumstances, then it is a non-trivial task demand relative to that system. This problem is exemplified by a recent debate about analogical reasoning. \citet{webbEmergentAnalogicalReasoning2023} showed that LLMs can match or surpass average human performance on various novel analogical reasoning tasks, including letter-string analogies such as $[ABC] \rightarrow [ABE], [MNO] \rightarrow [?]$. However, \citet{lewisUsingCounterfactualTasks2024} found that performance deteriorates when using a variant with a permuted alphabet. They interpret this drop as evidence that LLMs lack general competence in analogical reasoning. In response, \citet{webbEvidenceCounterfactualTasks2024} argue that solving the permuted task requires counting letter indices, which LLMs cannot do effectively without tools like a Python interpreter \citep[see][]{changLanguageModelsNeed2024}. Counting, they contend, imposes an auxiliary task demand that undermines the inference from poor performance to a lack of analogical competence. While humans may not assign numerical indices explicitly, they still must track \textit{how many times} to apply successor or predecessor operations. The permuted alphabet task requires counting positions relative to the provided key---an auxiliary demand that is trivial for a human with the key in view but difficult for an LLM that cannot reliably count without tools. This asymmetry in the impact of counting-related demands could explain the performance gap on counterfactual letter-string tasks.

Another kind of auxiliary factor is what we call \textit{test-time computational bottlenecks}—variable constraints on the computations a model can perform at inference time. This is distinct from an auxiliary task demand, which imposes an additional, conceptually separate task. A computational bottleneck, by contrast, is a resource limitation on the primary task itself.  This is most apparent in the context of \textit{test-time scaling}, a family of techniques for dynamically allocating compute during inference. The simplest example involves prompting a language model to generate a ``chain of thought'' (CoT) before answering \citep{weiChainofThoughtPromptingElicits2023}. A helpful analogy is the difference between solving a math problem with and without pen and paper. Prompting an LLM for a direct answer is loosely analogous to asking for a mental calculation, while allowing it to generate a CoT is loosely analogous to providing pen and paper for intermediate steps. When asked for a direct answer, Transformer models are fundamentally limited; theoretical work shows they cannot solve many sequential reasoning problems in a single forward pass \citep{merrillExpressivePowerTransformers2024}. Prompting them to ``think step by step'' triggers the generation of intermediate tokens that scaffold reasoning and substantially improve performance \citep{kojimaLargeLanguageModels2023}. In theory, each additional decoding step expands the model’s expressive power \citep{merrillExpressivePowerTransformers2024}. In practice, mechanistic analyses confirm that models use CoT tokens actively, drawing on multiple circuits to extract information from intermediate steps and assemble a final answer \citep{duttaHowThinkStepbystep2024}. The stark difference in performance with and without CoT shows that latent capabilities may go unexpressed when models are prompted to answer directly.

The influence of test-time computational bottlenecks is demonstrated more starkly by ``large reasoning models'' (LRMs), such as OpenAI o1 or DeepSeek R1, which are specifically trained to benefit from test-time scaling. RLMs are language models that undergo an additional post-training stage, which typically includes supervised fine-tuning on curated examples of CoT reasoning, as well as reinforcement learning on verifiable rewards in formal domains. At inference time, the computational budget for these models can be dynamically controlled by modulating the number of ``thinking'' tokens they generate before providing an answer. For example, \citet{muennighoffS1SimpleTesttime2025} found that increasing the ``thinking'' token budget would increase performance on difficult math and science benchmarks. In this case, it would be premature to conclude that performance failures with a low ``thinking'' token budget shows that the model lacks competence in the task domain. For the very same model and task, performance can switch from failure to success simply by letting the model generate a longer CoT.

The notion of computational bottlenecks extends beyond the generation of external intermediate tokens to scaffold computations. Experimental language models designed with an intrinsic capacity for test-time scaling \textit{in latent space} provide an even clearer illustration of the phenomenon. For example, \citet{geipingScalingTestTimeCompute2025} introduce a recurrent architecture that allows the model to ``think'' by iteratively refining its internal state in a continuous latent space, without ever needing to verbalize intermediate steps. They found that their model's performance on reasoning benchmarks improved when they increased the number of latent iterations at test time. For the very same problem, the model might fail with a few latent iterations but succeed with dozens. This demonstrates a purely computational bottleneck, disentangled from the auxiliary demand of articulating a chain of thought in natural language. To illustrate this phenomenon, we ran the latent reasoning model from \citet{geipingScalingTestTimeCompute2025} on examples from the Winogrande commonsense reasoning benchmark. The task involves filling in the blank in sentences like ``I tried to make mini lamps by using glow sticks in mason jars, but had to get larger jars because the \_\_\_ were too big.'' Initially, with only a few latent steps, the model assigns a higher log-likelihood to the incorrect completion, ``jars''. However, as the number of latent steps increases, the log-likelihood for the correct completion, ``glow sticks'', increases to become the dominant prediction.\footnote{Figure available at \url{https://github.com/raphael-milliere/latent-test-time-scaling}.} The model's initial failure is not evidence of its inability to solve the problem, but rather the consequence of a task-independent computational bottleneck. Its eventual success reveals a latent competence that is only expressed when sufficient computational resources are allocated at test time.

Finally, a third kind of auxiliary factor is what we might call \textit{mechanistic interference}, according to which a network learns a mechanism for solving a certain kind of problem, but the activity of that mechanism is interrupted by another, conceptually distinct process. Recent work on Transformer models trained on modular addition tasks provides insights into how a model's performance can be impacted by such auxiliary factors. \citet{nandaProgressMeasuresGrokking2022} show that even after a Transformer has formed a generalizable circuit that implements an algorithm to solve modular addition, its test set performance can still be impeded by previously memorized input-output mappings until a later ``cleanup'' phase in which this rote memorization is removed. This shows that the presence of a general algorithm does not guarantee strong test set performance when there is interference from auxiliary computations. \citet{zhongClockPizzaTwo2023} further show that even in fully trained networks that have ``grokked'' the task, there can be multiple concurrent circuits that interact to solve the task and potentially interfere with each other. For instance, a Transformer might contain one circuit implementing a general algorithm alongside another implementing a more heuristic algorithm for the same task. The model's overall performance could then be negatively impacted by the causal interference of the lesser algorithm, even though the latent competence to solve the task is present. 

The distinction between mechanistic interference and auxiliary task demand lies in the source of the failure: a task demand is an extrinsic feature of the experimental setup, while mechanistic interference is an intrinsic property of the model's internal processing, where two or more circuits produce conflicting outputs. One might object that this is an overly charitable interpretation; perhaps the interference simply reveals that the model's competence is fragile. However, the mechanistic separability of the competent circuit from the interfering process suggests otherwise. The point is not that the model is infallible, but that a general, competent mechanism exists and its function can be isolated, even if its output is sometimes suppressed. These findings illustrate that a model's performance on a task can be disrupted by auxiliary computations that are conceptually distinct from the core competence required for the task.

\subsection{Mechanistic chauvinism}

Mechanistic chauvinism is the tendency to assume that even when LLMs match or exceed average human performance, any substantive difference in strategy counts as evidence that the model’s solution lacks generality. In slogan form: all cognitive kinds are \textit{human} cognitive kinds.\footnote{This bias resembles what \citet{bucknerMorganCanonMeet2013} calls ``anthropofabulation''—a combination of (i) human-centric definitions of psychological terms with (ii) an exaggeration of human cognitive abilities. Mechanistic chauvinism differs from anthropofabulation in that it focuses on process rather than semantics or performance. Nevertheless, both forms of bias risk setting standards that no non-human system could meet.} Whenever an LLM arrives at a solution through a different computational process than humans use, the mechanistic chauvinist concludes that the LLM lacks genuine competence, regardless of how well it performs. 

However, this line of thinking risks obscuring the real capabilities of LLMs. Consider a model that learns a general algorithm for addition, rather than simply memorizing a large set of specific addition problems and their solutions. Now suppose that humans use a somewhat different algorithm for addition. The mere fact that the LLM's approach differs from the human approach does not mean that the LLM lacks real competence at addition. What matters is whether the LLM has learned a robust, generalizable strategy. Indeed, we can easily imagine an LLM outperforming humans at addition while using a distinctly unhuman-like computational process.This concern is arguably reflected in ongoing debates about the linguistic capacities of LLMs. Current models are not only proficient at generating grammatically well-formed sentences, but can also match human performance on judgments of syntactic acceptability, often aligning closely with human intuitions \citep{huLanguageModelsAlign2024a}. Yet, many linguists in the generative tradition, would argue that LLMs lack genuine syntactic competence \citep{chomskyAspectsTheorySyntax1965,chomskyNoamChomskyFalse2023,everaertStructuresNotStrings2015}. From this perspective, their success stems not from an internal knowledge of hierarchical syntactic rules, but from their nature as massively scaled statistical models trained to predict the next word. This conclusion, however, is challenged by a growing body of mechanistic evidence suggesting that LLMs do learn and represent hierarchical syntax internally, even if the computations that produce these representations differ from those posited in formal linguistic theories \citep{milliereLanguageModelsModelsforthcoming}. This debate thus exemplifies mechanistic chauvinism: if competence is defined by the instantiation of a specific computational architecture hypothesized in humans, then LLMs are deemed incompetent by fiat---thereby dismissing both their behavioral success and the alternative, learned mechanisms that support it.

\subsection{An objection}

We now anticipate an objection to our discussion of mechanistic chauvinism: since the only indisputable examples of language-driven cognition are human, and since the capacities of LLMs are acquired through training on human-produced data, isn't human cognition the most appropriate yardstick for studying LLMs? Our response is that it depends on how the yardstick is used. While we can -- and presumably must -- \textit{begin} our investigation of LLM competencies by comparing them to our own, the role of human cognition as a benchmark can be overstated in two ways. First, the human competencies we use as reference points should be regarded as what philosopher Ali Boyle calls ``investigative kinds'' \citep{boyleDisagreementClassificationComparative2024} - they serve as an initial search template, but not as necessary conditions for cognitive status. Second, questions of the form ``Do LLMs have cognitive competence $C$?'' should be treated as \textit{empirical} questions. Though this may seem self-evident, it is a principle that is easy to violate, particularly when we assume that facts about implementation are among the distinguishing features between competence $A$ and competence $B$. To preserve the empirical character of debates about the capacities of LLMs, we must focus on the appropriate level of analysis. Following Marr's classic framework, this means focusing on the computational and algorithmic levels, where we ask both \textit{what} the system computes and \textit{which} algorithms enable it to perform that computation. Defining competence at the physical or implementational level---for example, by requiring a biological substrate---would make the question of whether an LLM could possess a human-like competence non-empirical by definition. Therefore, our investigation explicitly sets aside facts about physical implementation.

The investigation of cognitive capacities in LLMs is best viewed as an iterative process in which our conception of LLM competencies, and of the mechanisms that implement them, mutually inform and revise each other. Solving the challenge of mapping cognitive tasks to capacities often involves consideration of the underlying mechanisms \citep{franckenCognitiveOntologySearch2022}. But to home in on the mechanisms responsible for a particular competence, we must make principled decisions about the level of abstraction at which to characterize the mechanism and about how to delineate the boundaries of the mechanism itself. These decisions, in turn, depend on how we conceptualize the cognitive phenomenon we are trying to explain. What results is an investigative process in which our characterization of cognitive tasks, our ontology of capacities, and our understanding of mechanisms evolve in tandem. In the case of LLMs, this process may lead us quite far from our initial starting point, and may ultimately produce a novel ontology of cognitive kinds, optimized for explaining the distinctive strengths and weaknesses of machine intelligence rather than human intelligence. In this way, the iterative nature of the investigative process allows it to drift away from its anthropocentric origins.    

\section{Conclusion}

Like humans, LLMs can be right for the wrong reasons. However, they can also be wrong for the wrong reasons. Auxiliary factors can suppress performance in ways that obscure latent competence. While anthropomorphism has received substantial critical attention, anthropocentric bias has gone largely unexamined. We have argued that it poses a serious obstacle to the objective assessment of LLM capacities, and we offered a taxonomy of its forms to help guide future research.

While carefully designed behavioral experiments are the primary means of identifying performance failures, mechanistic interpretability techniques are uniquely suited to adjudicate between competing hypotheses about the causes of these failures. For instance, when behavioral evidence is ambiguous, intepretability methods can, in principle, help distinguish whether a failure is due to a lack of a competent circuit or due to an auxiliary factor (such as mechanistic interference) suppressing an existing competent circuit. As such, supplementing behavioral with mechanistic evidence is a useful way to counteract auxiliary oversight. 

The role of mechanistic interpretability in addressing mechanistic chauvinism is more nuanced. It is possible to decode a feature or circuit of interest from a large neural network even if that feature or circuit does not significantly contribute to the network's functionality \citep{makelovThisSubspaceYou2024,huangRigorouslyAssessingNatural2023}. So there is a risk of anthropomorphic projection in mechanistic interpretability research. Nevertheless, there are examples of robust information processing mechanisms that differ from the ones humans typically use. Mechanistic interpretability research can help identify these mechanisms. When they are discovered, we should not dismiss them simply because they deviate from the human case. Doing so would foreclose the possibility of understanding artificial cognition on its own terms.

\bibliography{bibliography}

@book{chomskyAspectsTheorySyntax1965,
	title        = {Aspects of the {{Theory}} of {{Syntax}}},
	author       = {Chomsky, Noam},
	year         = 1965,
	publisher    = {Cambridge, MA, USA: MIT Press}
}

@article{chomskyNoamChomskyFalse2023,
	title        = {Noam {{Chomsky}}: {{The False Promise}} of {{ChatGPT}}},
	shorttitle   = {Opinion {\textbar} {{Noam Chomsky}}},
	author       = {Chomsky, Noam and Roberts, Ian and Watumull, Jeffrey},
	year         = 2023,
	month        = mar,
	journal      = {The New York Times},
	issn         = {0362-4331},
	urldate      = {2023-08-20},
	chapter      = {Opinion},
	langid       = {american}
}

@article{everaertStructuresNotStrings2015,
	title        = {Structures, {{Not Strings}}: {{Linguistics}} as {{Part}} of the {{Cognitive Sciences}}},
	shorttitle   = {Structures, {{Not Strings}}},
	author       = {Everaert, Martin B. H. and Huybregts, Marinus A. C. and Chomsky, Noam and Berwick, Robert C. and Bolhuis, Johan J.},
	year         = 2015,
	month        = dec,
	journal      = {Trends in Cognitive Sciences},
	volume       = 19,
	number       = 12,
	pages        = {729--743},
	doi          = {10.1016/j.tics.2015.09.008},
	issn         = {1364-6613},
	urldate      = {2023-07-30},
	langid       = {english}
}

@article{huLanguageModelsAlign2024a,
	title        = {Language Models Align with Human Judgments on Key Grammatical Constructions},
	author       = {Hu, Jennifer and Mahowald, Kyle and Lupyan, Gary and Ivanova, Anna and Levy, Roger},
	year         = 2024,
	month        = sep,
	journal      = {Proceedings of the National Academy of Sciences},
	publisher    = {Proceedings of the National Academy of Sciences},
	volume       = 121,
	number       = 36,
	pages        = {e2400917121},
	doi          = {10.1073/pnas.2400917121},
	urldate      = {2025-10-16},
	langid       = {english}
}

@article{ivanovaHowEvaluateCognitive2025,
	title        = {How to Evaluate the Cognitive Abilities of {{LLMs}}},
	author       = {Ivanova, Anna A.},
	year         = 2025,
	month        = feb,
	journal      = {Nature Human Behaviour},
	publisher    = {Nature Publishing Group},
	volume       = 9,
	number       = 2,
	pages        = {230--233},
	doi          = {10.1038/s41562-024-02096-z},
	issn         = {2397-3374},
	urldate      = {2025-10-16},
	langid       = {english}
}

@incollection{milliereLanguageModelsModelsforthcoming,
	title        = {Language {{Models}} as {{Models}} of {{Language}}},
	author       = {Milli{\`e}re, Rapha{\"e}l},
	year         = {forthcoming},
	booktitle    = {The {{Oxford Handbook}} of the {{Philosophy}} of {{Linguistics}}},
	publisher    = {Oxford University Press},
	address      = {Oxford},
	editor       = {Nefdt, Ryan and Dupre, Gabe and Stanton, Kate}
}

@misc{benderNYTMagazineAI2022,
	title        = {On {{NYT Magazine}} on {{AI}}: {{Resist}} the {{Urge}} to Be {{Impressed}}},
	shorttitle   = {On {{NYT Magazine}} on {{AI}}},
	author       = {Bender, Emily M.},
	year         = 2022,
	month        = may,
	journal      = {Medium},
	urldate      = {2024-07-04}
}

@article{boyleDisagreementClassificationComparative2024,
	title        = {Disagreement \& Classification in Comparative Cognitive Science},
	author       = {Boyle, Alexandria},
	year         = 2024,
	journal      = {No{\^u}s},
	volume       = {n/a},
	number       = {n/a},
	doi          = {10.1111/nous.12480},
	issn         = {1468-0068},
	urldate      = {2024-03-05}
}

@article{bucknerBlackBoxesUnflattering2021,
	title        = {Black {{Boxes}} or {{Unflattering Mirrors}}? {{Comparative Bias}} in the {{Science}} of {{Machine Behaviour}}},
	shorttitle   = {Black {{Boxes}} or {{Unflattering Mirrors}}?},
	author       = {Buckner, Cameron},
	year         = 2021,
	month        = apr,
	journal      = {The British Journal for the Philosophy of Science},
	publisher    = {The University of Chicago Press},
	pages        = {000--000},
	doi          = {10.1086/714960},
	issn         = {0007-0882},
	urldate      = {2023-08-09}
}

@article{bucknerMorganCanonMeet2013,
	title        = {Morgan's {{Canon}}, Meet {{Hume}}'s {{Dictum}}: Avoiding Anthropofabulation in Cross-Species Comparisons},
	shorttitle   = {Morgan's {{Canon}}, Meet {{Hume}}'s {{Dictum}}},
	author       = {Buckner, Cameron},
	year         = 2013,
	month        = sep,
	journal      = {Biology \& Philosophy},
	volume       = 28,
	number       = 5,
	pages        = {853--871},
	doi          = {10.1007/s10539-013-9376-0},
	issn         = {1572-8404},
	urldate      = {2023-08-09}
}

@misc{changLanguageModelsNeed2024,
	title        = {Language {{Models Need Inductive Biases}} to {{Count Inductively}}},
	author       = {Chang, Yingshan and Bisk, Yonatan},
	year         = 2024,
	month        = may,
	publisher    = {arXiv},
	number       = {arXiv:2405.20131},
	doi          = {10.48550/arXiv.2405.20131},
	urldate      = {2024-06-07},
	eprint       = {2405.20131},
	archiveprefix = {arXiv}
}

@misc{duttaHowThinkStepbystep2024,
	title        = {How to Think Step-by-Step: {{A}} Mechanistic Understanding of Chain-of-Thought Reasoning},
	shorttitle   = {How to Think Step-by-Step},
	author       = {Dutta, Subhabrata and Singh, Joykirat and Chakrabarti, Soumen and Chakraborty, Tanmoy},
	year         = 2024,
	month        = may,
	publisher    = {arXiv},
	number       = {arXiv:2402.18312},
	doi          = {10.48550/arXiv.2402.18312},
	urldate      = {2025-06-11},
	eprint       = {2402.18312},
	archiveprefix = {arXiv}
}

@article{firestonePerformanceVsCompetence2020,
	title        = {Performance vs. Competence in Human--Machine Comparisons},
	author       = {Firestone, Chaz},
	year         = 2020,
	month        = oct,
	journal      = {Proceedings of the National Academy of Sciences},
	publisher    = {National Academy of Sciences},
	volume       = 117,
	number       = 43,
	pages        = {26562--26571},
	doi          = {10.1073/pnas.1905334117},
	isbn         = 9781905334117,
	issn         = {0027-8424, 1091-6490},
	urldate      = {2021-03-15},
	chapter      = {Perspective},
	pmid         = 33051296
}

@article{franckenCognitiveOntologySearch2022,
	title        = {Cognitive Ontology and the Search for Neural Mechanisms: Three Foundational Problems},
	shorttitle   = {Cognitive Ontology and the Search for Neural Mechanisms},
	author       = {Francken, Jolien C. and Slors, Marc and Craver, Carl F.},
	year         = 2022,
	month        = sep,
	journal      = {Synthese},
	volume       = 200,
	number       = 5,
	pages        = 378,
	doi          = {10.1007/s11229-022-03701-2},
	issn         = {1573-0964},
	urldate      = {2024-03-05}
}

@misc{geipingScalingTestTimeCompute2025,
	title        = {Scaling up {{Test-Time Compute}} with {{Latent Reasoning}}: {{A Recurrent Depth Approach}}},
	shorttitle   = {Scaling up {{Test-Time Compute}} with {{Latent Reasoning}}},
	author       = {Geiping, Jonas and McLeish, Sean and Jain, Neel and Kirchenbauer, John and Singh, Siddharth and Bartoldson, Brian R. and Kailkhura, Bhavya and Bhatele, Abhinav and Goldstein, Tom},
	year         = 2025,
	month        = feb,
	publisher    = {arXiv},
	number       = {arXiv:2502.05171},
	doi          = {10.48550/arXiv.2502.05171},
	urldate      = {2025-02-11},
	eprint       = {2502.05171},
	archiveprefix = {arXiv}
}

@misc{huangRigorouslyAssessingNatural2023,
	title        = {Rigorously {{Assessing Natural Language Explanations}} of {{Neurons}}},
	author       = {Huang, Jing and Geiger, Atticus and D'Oosterlinck, Karel and Wu, Zhengxuan and Potts, Christopher},
	year         = 2023,
	month        = sep,
	publisher    = {arXiv},
	urldate      = {2024-01-11}
}

@misc{huAuxiliaryTaskDemands2024,
	title        = {Auxiliary Task Demands Mask the Capabilities of Smaller Language Models},
	author       = {Hu, Jennifer and Frank, Michael C.},
	year         = 2024,
	month        = apr,
	publisher    = {arXiv},
	number       = {arXiv:2404.02418},
	doi          = {10.48550/arXiv.2404.02418},
	urldate      = {2024-04-06},
	eprint       = {2404.02418},
	archiveprefix = {arXiv}
}

@misc{kojimaLargeLanguageModels2023,
	title        = {Large {{Language Models}} Are {{Zero-Shot Reasoners}}},
	author       = {Kojima, Takeshi and Gu, Shixiang Shane and Reid, Machel and Matsuo, Yutaka and Iwasawa, Yusuke},
	year         = 2023,
	month        = jan,
	publisher    = {arXiv},
	number       = {arXiv:2205.11916},
	doi          = {10.48550/arXiv.2205.11916},
	urldate      = {2025-06-11},
	eprint       = {2205.11916},
	archiveprefix = {arXiv}
}

@inproceedings{lakretzCanTransformersProcess2022,
	title        = {Can {{Transformers Process Recursive Nested Constructions}}, {{Like Humans}}?},
	author       = {Lakretz, Yair and Desbordes, Th{\'e}o and Hupkes, Dieuwke and Dehaene, Stanislas},
	year         = 2022,
	month        = oct,
	booktitle    = {Proceedings of the 29th {{International Conference}} on {{Computational Linguistics}}},
	publisher    = {International Committee on Computational Linguistics},
	address      = {Gyeongju, Republic of Korea},
	pages        = {3226--3232},
	urldate      = {2023-08-08}
}

@misc{lampinenCanLanguageModels2023,
	title        = {Can Language Models Handle Recursively Nested Grammatical Structures? {{A}} Case Study on Comparing Models and Humans},
	shorttitle   = {Can Language Models Handle Recursively Nested Grammatical Structures?},
	author       = {Lampinen, Andrew Kyle},
	year         = 2023,
	month        = feb,
	publisher    = {arXiv},
	number       = {arXiv:2210.15303},
	doi          = {10.48550/arXiv.2210.15303},
	urldate      = {2023-08-08},
	eprint       = {2210.15303},
	archiveprefix = {arXiv}
}

@misc{lewisUsingCounterfactualTasks2024,
	title        = {Using {{Counterfactual Tasks}} to {{Evaluate}} the {{Generality}} of {{Analogical Reasoning}} in {{Large Language Models}}},
	author       = {Lewis, Martha and Mitchell, Melanie},
	year         = 2024,
	month        = feb,
	publisher    = {arXiv},
	number       = {arXiv:2402.08955},
	doi          = {10.48550/arXiv.2402.08955},
	urldate      = {2024-02-18},
	eprint       = {2402.08955},
	archiveprefix = {arXiv}
}

@inproceedings{makelovThisSubspaceYou2024,
	title        = {Is {{This}} the {{Subspace You Are Looking}} for? {{An Interpretability Illusion}} for {{Subspace Activation Patching}}},
	shorttitle   = {Is {{This}} the {{Subspace You Are Looking}} For?},
	author       = {Makelov, Aleksandar and Lange, Georg and Geiger, Atticus and Nanda, Neel},
	year         = 2024,
	booktitle    = {The {{Twelfth International Conference}} on {{Learning Representations}}, {{ICLR}} 2024, {{Vienna}}, {{Austria}}, {{May}} 7-11, 2024},
	publisher    = {OpenReview.net},
	urldate      = {2025-03-12}
}

@misc{merrillExpressivePowerTransformers2024,
	title        = {The {{Expressive Power}} of {{Transformers}} with {{Chain}} of {{Thought}}},
	author       = {Merrill, William and Sabharwal, Ashish},
	year         = 2024,
	month        = mar,
	publisher    = {arXiv},
	number       = {arXiv:2310.07923},
	doi          = {10.48550/arXiv.2310.07923},
	urldate      = {2024-04-01},
	eprint       = {2310.07923},
	archiveprefix = {arXiv}
}

@misc{muennighoffS1SimpleTesttime2025,
	title        = {S1: {{Simple}} Test-Time Scaling},
	shorttitle   = {S1},
	author       = {Muennighoff, Niklas and Yang, Zitong and Shi, Weijia and Li, Xiang Lisa and {Fei-Fei}, Li and Hajishirzi, Hannaneh and Zettlemoyer, Luke and Liang, Percy and Cand{\`e}s, Emmanuel and Hashimoto, Tatsunori},
	year         = 2025,
	month        = mar,
	publisher    = {arXiv},
	number       = {arXiv:2501.19393},
	doi          = {10.48550/arXiv.2501.19393},
	urldate      = {2025-05-19},
	eprint       = {2501.19393},
	archiveprefix = {arXiv}
}

@inproceedings{nandaProgressMeasuresGrokking2022,
	title        = {Progress Measures for Grokking via Mechanistic Interpretability},
	author       = {Nanda, Neel and Chan, Lawrence and Lieberum, Tom and Smith, Jess and Steinhardt, Jacob},
	year         = 2022,
	month        = sep,
	booktitle    = {The {{Eleventh International Conference}} on {{Learning Representations}}},
	urldate      = {2023-08-11}
}

@incollection{soberComparativePsychologyMeets2005,
	title        = {Comparative Psychology Meets Evolutionary Biology},
	author       = {Sober, Elliott},
	year         = 2005,
	booktitle    = {Thinking with Animals: {{New}} Perspectives on Anthropomorphism},
	publisher    = {Columbia University Press},
	urldate      = {2024-05-31},
	editor       = {Datson, Lorraine and Mitman, Gregg}
}

@article{webbEmergentAnalogicalReasoning2023,
	title        = {Emergent Analogical Reasoning in Large Language Models},
	author       = {Webb, Taylor and Holyoak, Keith J. and Lu, Hongjing},
	year         = 2023,
	month        = jul,
	journal      = {Nature Human Behaviour},
	publisher    = {Nature Publishing Group},
	pages        = {1--16},
	doi          = {10.1038/s41562-023-01659-w},
	issn         = {2397-3374},
	urldate      = {2023-08-29}
}

@misc{webbEvidenceCounterfactualTasks2024,
	title        = {Evidence from Counterfactual Tasks Supports Emergent Analogical Reasoning in Large Language Models},
	author       = {Webb, Taylor and Holyoak, Keith J. and Lu, Hongjing},
	year         = 2024,
	month        = apr,
	journal      = {arXiv.org},
	urldate      = {2024-04-26},
	howpublished = {https://arxiv.org/abs/2404.13070v1}
}

@misc{weiChainofThoughtPromptingElicits2023,
	title        = {Chain-of-{{Thought Prompting Elicits Reasoning}} in {{Large Language Models}}},
	author       = {Wei, Jason and Wang, Xuezhi and Schuurmans, Dale and Bosma, Maarten and Ichter, Brian and Xia, Fei and Chi, Ed and Le, Quoc and Zhou, Denny},
	year         = 2023,
	month        = jan,
	publisher    = {arXiv},
	number       = {arXiv:2201.11903},
	doi          = {10.48550/arXiv.2201.11903},
	urldate      = {2025-05-19},
	eprint       = {2201.11903},
	archiveprefix = {arXiv}
}

@misc{zhongClockPizzaTwo2023,
	title        = {The {{Clock}} and the {{Pizza}}: {{Two Stories}} in {{Mechanistic Explanation}} of {{Neural Networks}}},
	shorttitle   = {The {{Clock}} and the {{Pizza}}},
	author       = {Zhong, Ziqian and Liu, Ziming and Tegmark, Max and Andreas, Jacob},
	year         = 2023,
	month        = jun,
	publisher    = {arXiv},
	number       = {arXiv:2306.17844},
	doi          = {10.48550/arXiv.2306.17844},
	urldate      = {2023-08-11},
	eprint       = {2306.17844},
	archiveprefix = {arXiv}
}
\end{document}